%% file: acl_latex.tex
\pdfoutput=1

\documentclass[11pt]{article}

\usepackage[preprint]{acl}

\usepackage{times}
\usepackage{latexsym}

\usepackage[T1]{fontenc}

\usepackage{listings}
\usepackage[utf8]{inputenc}

\usepackage{microtype}

\usepackage{inconsolata}

\usepackage{times}
\usepackage{latexsym}
\usepackage[T1]{fontenc}
\usepackage[utf8]{inputenc}
\usepackage{microtype}
\usepackage{inconsolata}
\usepackage{graphicx}
\usepackage{sidecap}
\sidecaptionvpos{figure}{c} 

\usepackage{amsmath}
\usepackage{adjustbox}
\usepackage{microtype}
\usepackage{hyperref}
\usepackage{url}
\usepackage{booktabs}
\usepackage{geometry}
\usepackage{multirow}
\usepackage{algorithm}
\usepackage{algpseudocode}
\usepackage{wrapfig}
\usepackage{subcaption}
\usepackage{amsmath}
\usepackage{listings}
\usepackage{natbib}
\usepackage{bbm}
\usepackage{float}
\usepackage{xcolor}
\usepackage{caption}
\usepackage{makecell}

\usepackage{xspace}

\newcommand\eg{e.g.,~}
\newcommand\ie{i.e.,~}

\newcommand{\name}{L2R\xspace}
\newcommand{\base}{RAIDAR\xspace}
\usepackage{enumitem}
\newenvironment{denseitemize}{
\begin{itemize}[topsep=2pt, partopsep=0pt, leftmargin=2em]
  \setlength{\itemsep}{2pt}
  \setlength{\parskip}{0pt}
  \setlength{\parsep}{0pt}
}{\end{itemize}}

\newcommand{\revision}[1]{\textcolor{black}{#1}}

\input{def}

\title{Learning to Rewrite: Generalized LLM-Generated Text Detection}
\author{Ran Li$^{1}$\thanks{Equal contribution}, \ \ Wei Hao$^{1}$\footnotemark[1], \ \ Weiliang Zhao$^{1}$, \ \ Junfeng Yang$^{1}$, \ \ Chengzhi Mao$^{2}$ \\
Columbia University$^{1}$, Rutgers University$^{2}$\\
\texttt{\{rl3424, wh2473, wz2665, jy2324\}@columbia.edu, cm1838@rutgers.edu}
}

\begin{document}
\maketitle

\input{latex/abstract}
\input{latex/introduction}
\input{latex/related_work}
\input{latex/method}

\input{latex/dataset}
\input{latex/evaluation}
\input{latex/conclusion}
\input{latex/limitations}

\bibliography{custom}

\appendix

\input{latex/appendix}

\end{document}

%% file: def.tex
\def\X{{\bf X}}
\def\Y{{\bf Y}}

\def\x{{\bf x}}

\def\0{{\bf 0}}
\def\1{{\bf 1}}

\def\id{ID\xspace}
\def\ood{OOD\xspace}

%% file: latex/abstract.tex
\begin{abstract}
Large language models (LLMs) present significant risks when used to generate non-factual content and spread disinformation at scale. Detecting such LLM-generated content is crucial, yet current detectors often struggle to generalize in open-world contexts. We introduce \textbf{Learning2Rewrite}, a novel framework for detecting AI-generated text with exceptional generalization to unseen domains. Our method leverages the insight that LLMs inherently modify AI-generated content less than human-written text when tasked with rewriting. By training LLMs to minimize alterations on AI-generated inputs, we amplify this disparity, yielding a more distinguishable and generalizable edit distance across diverse text distributions. Extensive experiments on data from 21 independent domains and four major LLMs (GPT-3.5, GPT-4, Gemini, and Llama-3) demonstrate that our detector outperforms state-of-the-art detection methods by up to 23.04\% in AUROC for in-distribution tests, 37.26\% for out-of-distribution tests, and 48.66\% under adversarial attacks. Our unique training objective ensures better generalizability compared to directly training for classification, when leveraging the same amount of parameters. Our findings suggest that reinforcing LLMs’ inherent rewriting tendencies offers a robust and scalable solution for detecting AI-generated text.
\end{abstract}

%% file: latex/introduction.tex
\section{Introduction}
Large Language Models (LLMs) demonstrate exceptional capabilities across various tasks~\citep{radford2019language, brown2020language, achiam2023gpt, touvron2023Llama, team2023gemini, openai_chatgpt}. However, they can be misused for illegal or unethical activities, such as spreading misinformation~\citep{chen2023combating}, scaling spear phishing campaigns~\citep{hazell2023large}, facilitating social engineering and manipulation of social media~\citep{zhang2024toward}, and generating propaganda~\citep{pan2023risk}. LLMs also facilitate academic dishonesty~\citep{zellers2019defending, mvondo2023generative}, and training foundation models with generated content can lead to irreversible defects in resulting models~\citep{shumailov2023curse}. These issues highlight the urgent need for reliable algorithms to detect LLM-generated text.

\begin{figure}
  \centering
  \begin{subfigure}[b]{\columnwidth}
    \centering
    \includegraphics[width=0.32\columnwidth]{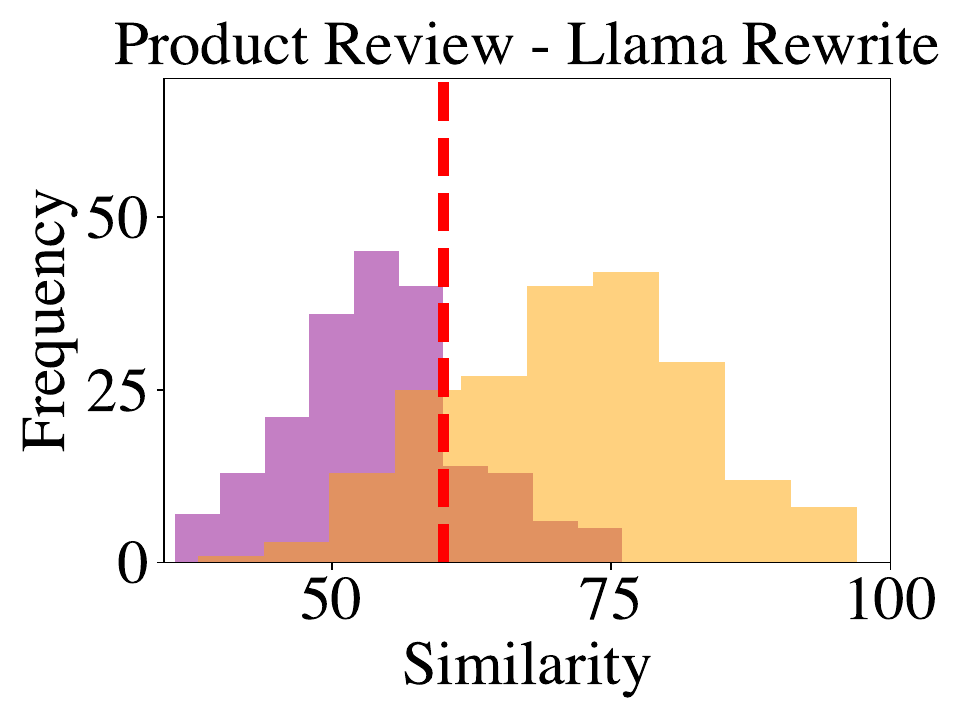}
    \includegraphics[width=0.32\columnwidth]{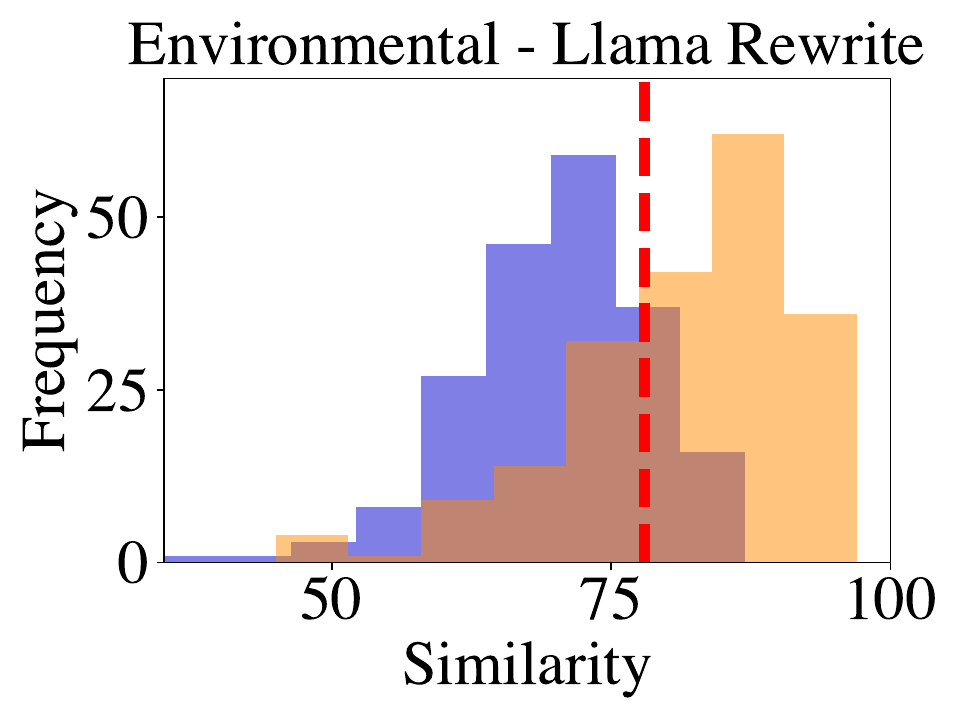}
    \includegraphics[width=0.28\columnwidth]{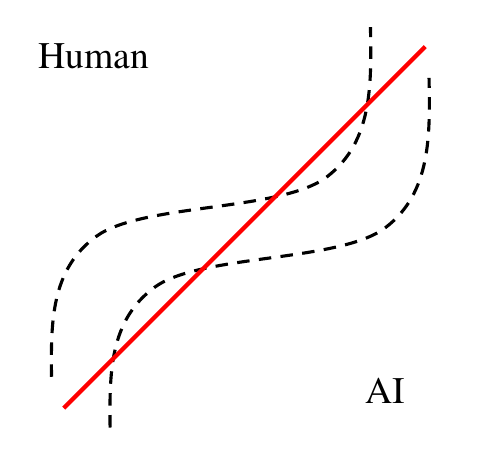}
  \end{subfigure}
  \begin{subfigure}[b]{\columnwidth}
    \centering
    \includegraphics[width=0.32\columnwidth]{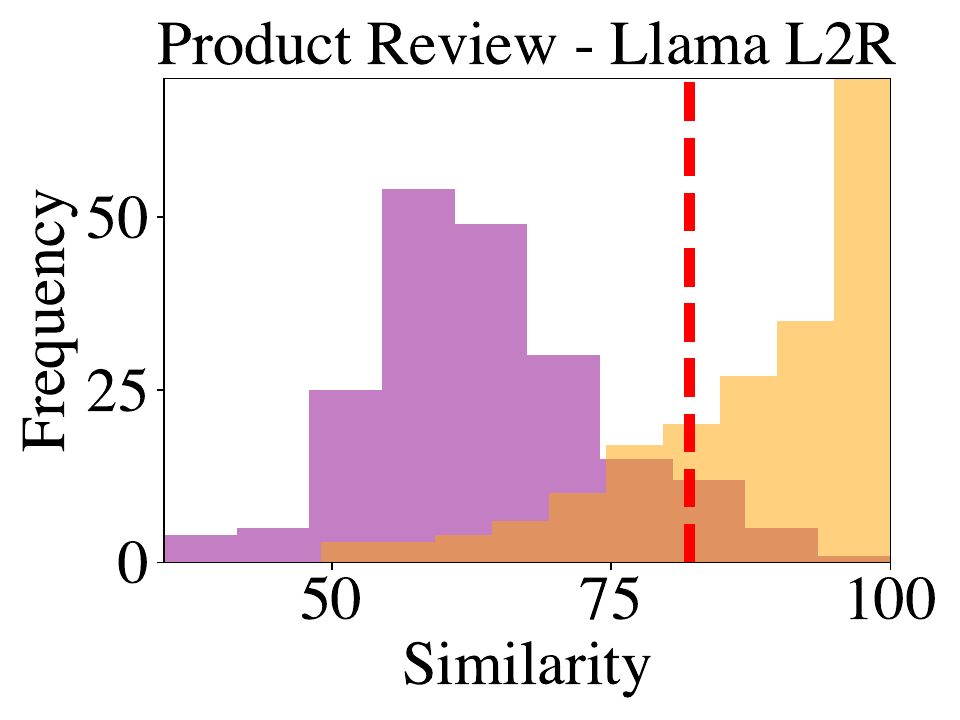}
    \includegraphics[width=0.32\columnwidth]{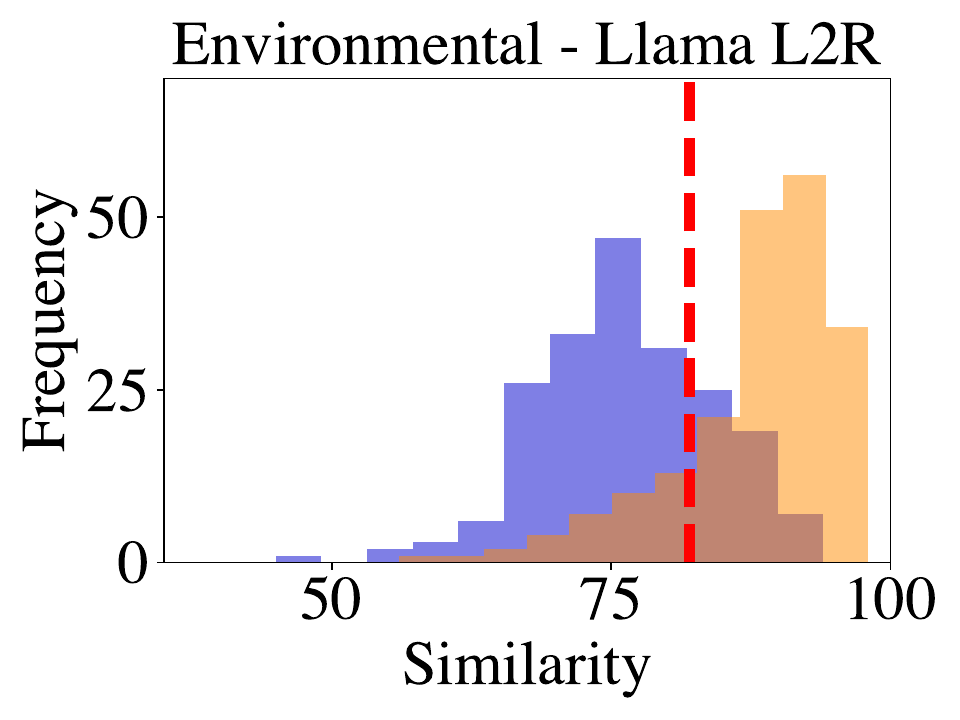}
    \includegraphics[width=0.28\columnwidth]{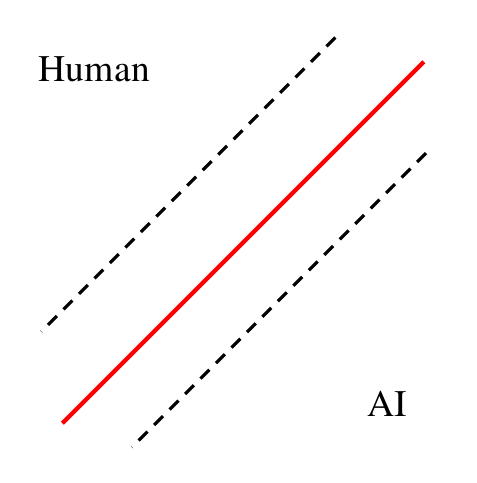}
  \end{subfigure}
  \caption{\textbf{Rewriting for LLM Text Detection}. The histograms depict the edit distance distributions for texts generated by human and AI, illustrating how fine-tuning a rewrite model enhances their separation. We show two domains: \textcolor{purple}{Purple} and \textcolor{yellow}{Yellow} represent human and AI distributions for Product Review texts, while \textcolor{blue}{Blue} and \textcolor{orange}{Orange} represent those for Environmental texts. Without fine-tuning the rewrite model, human and AI distributions are inseparable by a single threshold (red line, above). After fine-tuning, the texts can be separated by this threshold (below). On the right, we conceptualize \name's intuition by showing that the rugged decision boundary between human and AI texts, caused by varying data distributions across domains, can be better aligned and divided by a single threshold after fine-tuning. Specifically, the standard deviation in decision thresholds among all domains decreases from 0.7506 to 0.4428 after fine-tuning.}
  \label{fig:histogram}
\end{figure}

 Various methods for detecting generated text have been proposed~\citep{solaiman2019release,fagni2021tweepfake,mitrovic2023chatgpt, mitchell2023detectgpt,su2023detectllm, liu2024does, bao2023fast, mao2024raidar}.
Most of these detectors employ pre-trained models, extracting hand-crafted features and heuristics, such as loss curvature~\citep{bao2023fast} and rewriting distance~\citep{mao2024raidar}, and apply thresholds to distinguish LLM from human data. However, these thresholds are highly domain-dependent, obfuscating the establishment of a universal detection standard.

In this paper, we present \name (Learning to Rewrite), which trains an LLM to perform more edits when being asked to rewrite human-generated data and fewer edits when rewriting on LLM-generated data across a diverse set of domains. Unlike traditional detectors, which work well in-distribution (\id) but often struggle to generalize among out-of-distribution (\ood) domains (including adversarial attacks), our algorithm leverages the inherent tendency of LLMs to modify their own output less frequently, and maximizing its generalizability by focusing on learning a single rewriting threshold across diverse distributions. Figure~\ref{fig:histogram} illustrates an example of how \name learns to make LLM and human generated text more separable across domains, comparing with rewriting using a pre-trained model~\citep{mao2024raidar}.

Visualizations and numerical results demonstrate that our targeted training objective enables LLMs to better capture the intricate structure of AI-generated content. To reflect the rapid advancements and real-world diversity of LLM-generated text, we in addition constructed a dataset spanning 21 domains (e.g., finance, entertainment, cuisine) using four different generator models. 
\name surpasses the state-of-the-art detectors, achieving up to 19.56\% higher AUROC \id and 35.10\% higher \ood than~\citet{verma2023ghostbuster}, 23.04\% higher \id and 37.26\% higher \ood than~\citet{bao2023fast}, and 10.39\% higher \id and 4.67\% higher \ood than~\citet{mao2024raidar}. Comparing with fine-tuning a Llama-3 model for naive text classification, \name has 51.35\% higher AUROC \ood despite leveraging the same number of parameters. These results demonstrate that our training objective offers superior accuracy and generalizability. Furthermore, our method provides interpretability by highlighting the rewritten portions of the text. We will release our data, code, and models upon acceptance.

Our contributions are as follows:
\begin{itemize}
    \item Fine-tuned detectors for generated text detection are known for overfitting to specific domains. We propose L2R, whose learning objective is rather to enlarge the edit distance between rewriting and the original text for LLM-generated text while minimizing the ones that are human-generated. This learning objective is relatively domain-agnostic, yielding an invariant detection threshold across different data distributions.
    \item We build a diversely generated dataset (21 domains) and design a calibration loss function to make fine-tuning both effective and stable.
    \item We conduct comprehensive evaluations on ID, OOD datasets and against different adversarial attacks (Decoherence and Rewrite bypassing), showing that L2R surpasses state-of-the-art learning-based and zero-shot-based detectors.
\end{itemize}

\begin{figure*}[t]
    \centering
    \includegraphics[width=\textwidth]{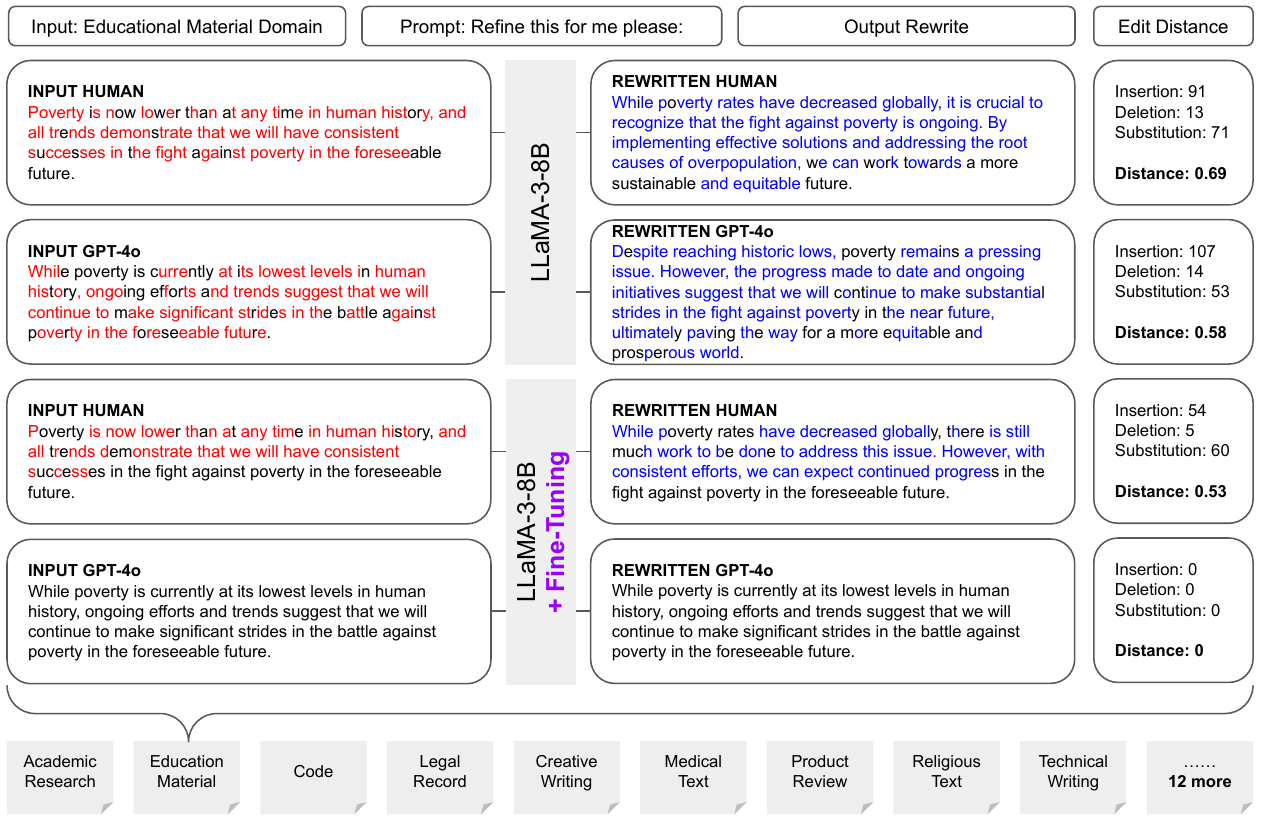}
    \caption{\textbf{Overview.} Deleted characters are marked in \textcolor{red}{red}, added characters are marked in \textcolor{blue}{blue}, and unmodified characters are in \textbf{black}. We exploit the difference in rewriting distance between human and AI texts for classification. While the off-the-shelf Llama-3 model give different amount of rewrite for human and AI texts (above), rewrites from our fine-tuned model are much more separable (below).}
    \label{fig:teaser}
\end{figure*}

%% file: latex/related_work.tex
\section{Related Work}
Various AI-generated text detectors have been proposed over the years. One set of detectors train a model on the input text~\citep{solaiman2019release,fagni2021tweepfake, shnarch2022cluster, mitrovic2023chatgpt, liu2023coco}. These methods excel in their training domains but struggle under \ood evaluation~\citep{uchendu-etal-2020-authorship, pu2023deepfake}, namely detection with text from different domains or unfamiliar models. The second set of detectors utilize the raw outputs, \ie logits, from pre-trained LLMs to assign probability score for detection. \revision{GLTR~\citep{gehrmann2019gltr} utilizes statistical features like log probability, word rank, and entropy to assign score,} Ghostbuster~\citep{verma2023ghostbuster} utilizes log probability and unigram and bigram probability, DetectGPT~\citep{mitchell2023detectgpt} employs the delta in log probability of the input text after token perturbation to estimate AI likehood, PECOLA~\cite{liu2024does} selectively applys perturbation for enhanced accuracy, and Fast-DetectGPT~\citep{bao2023fast} simplifies the process by exploiting conditional probability curvature. These family of detectors all require raw output of an LLM in some way or the other, but the main target of detection, namely commercial LLMs, are not open-sourced, which potentially impose a barrier on their probability estimation. Lastly, RAIDAR~\citep{mao2024raidar} is a detection method based on the observation that LLMs, when prompted to rewrite a given text, tend to produce a greater number of rewrites for human-written text compared to AI-generated text. Despite the attempt on capturing rewrite edit distance as a domain-agnostic feature, the rewrite amount still varies across distributions, \revision{and the threshold of rewrite amount between human and AI texts learned on training domains does not generalize to \ood,} which limits its full potential.

%% file: latex/method.tex
\section{Method}

\subsection{Rewriting for LLM Detection}

Rewriting input with LLM and then measuring the change proves to be a successful way to detect LLM-generated content. Given an held-out input text set $\X_{train}$ with LLM and human generated text, and its corresponding label set $\Y_{train}$, an LLM $F(\cdot)$ is prompted to rewrite the input $\x \in \X_{train}$ using a prompt ${\bf p}$. The rewriting output is $F({\bf p}, \x)$. Particularly, the prompt ${\bf p}$ can be set to: \texttt{Refine this for me please}.

The edit distance between the input text and the rewritten output, $D(\x, F({\bf p}, \x))$,  is then computed for all $\x \in \X_{train}$.~\citet{mao2024raidar} adopts the Levenshtein distance~\citep{levenshtein1966binary}, which is defined as the minimum number of insertions, deletions, or substitutions required to transform one text into the other. With the Levenshtein distance, a similarity score we use for classification is calculated based on:
\begin{equation}
    D_k(\x, F({\bf p}, \x)) = 1 - \frac{\text{Levenshtein}(F({\bf p}, \x), \x)}{\max(len(F({\bf p}, \x)), len(\x))}.
\end{equation}

\citet{mao2024raidar} trains a classifier, such as logistic regression or decision tree, to threshold the similarity scores and predict if it is written by an LLM. However, as shown in Figure~\ref{fig:histogram}, the threshold of rewriting with a vanilla LLM often varies from one domain to another, causing RAIDAR to fail to generalize to new domains.

\subsection{Fine-Tuning the Rewrite Model}
\name works on the premise that human-written and AI generated text would cause a different amount of rewrites and a boundary can be drawn to separate both distributions. Thus we can finetune such a rewrite model $F'(\cdot)$, that gives as much rewrite as possible for human texts, while leaving the AI texts unmodified, demonstrated in Figure~\ref{fig:teaser}. Given some human text $\x_h \in \X_{train}$ and AI text $\x_{ai} \in \X_{train}$, our objective becomes: 
\begin{equation}
\max\{D(\x_h, F'({\bf p}, \x_h)) - D(\x_{ai}, F'({\bf p}, \x_{ai}))\}
\end{equation}

Since the edit distance is not differentiable, we use the cross-entropy loss $L(\cdot)$ assigned to the input $\x$ by $F'(\cdot)$ as a proxy to the edit distance. As a result, for each of input $\x$ with label $y=1$ (AI) or $0$ (human), our loss function becomes:
\begin{equation}
\min \{ L(\mathbf{X}_{\text{train}}) \cdot y_{\text{train}} \}, \quad  y_{\text{train}} = 
\begin{cases} 
1 & \text{(AI)} \\ 
-1 & \text{(human)}
\end{cases}
\label{eq:proxy}
\end{equation}


In this way, we flip the sign of the loss of the human texts. Since the overall loss would be minimized, this effectively encourages the rewrites to be different from human input and identical to the AI input. 

\subsection{Calibration Loss during Fine-Tuning}
When fine-tuning the rewrite model on Equation~\ref{eq:proxy}, the rewrite model aims to maximize the edits on human-generated text and minimize the edits on LLM-generated texts. However, without posting regularization and constraint on the unbounded loss, the rewrite model takes the risk of being corrupted (\eg verbose output for all rewrite and over-fitting with more edits on human-generated text rewrite) which we evaluated in \S\ref{sec:loss_eval}.

Therefore, we propose a calibration loss, which prevents the over-fitting problem by imposing a threshold value $t$ on the absolute value of the loss on each given input. For human text $\x_h$, we apply gradient backpropagation only if the absolute loss $L(\x_h) < t$. For AI text $\x_{ai}$, we apply backpropagation only if $L(\x_{ai}) > t$. Otherwise, the gradient is set to 0. \revision{We show a pseudocode for the algorithm in \ref{alg:calibration}.}

\begin{algorithm}
\caption{Calibration Loss Calculation}
\begin{algorithmic}[1]
\Require Threshold $t$, loss $L(\cdot)$, human text $x_h$, AI text $x_{ai}$
\State $L_h \gets L(x_h)$, $L_{ai} \gets L(x_{ai})$, $L \gets 0$
\State $L \gets L + L_h \text{ if } L_h < t$
\State $L \gets L + L_{ai} \text{ if } L_{ai} > t$
\State \Return $L$
\end{algorithmic}
\label{alg:calibration}
\end{algorithm}


Therefore, rather than minimizing the loss proxy, our objective becomes separating the distribution of human and AI rewrites to two ends of the threshold $t$. \revision{Concretely, this enables the model to only optimize against the hard examples, and leave those already correctly classified unchanged, so that we prevent overfitting. This is similar to DPO \citep{rafailov2023direct}, where we fine-tune the rewrite model using only preference data, namely the rewrites that are not yet separated by the existing boundary.} This process is depicted by the graphical illustrations in Figure~\ref{fig:histogram}.

To determine the threshold $t$, we perform a forward pass using the rewrite model before fine-tuning on $\X_{train}$ and train a logistic regression model on all loss values. The threshold $t$ can be derived from the weight and the intercept of the logistic regression model. In practice, applying the calibration loss improves detection performance by 4.54\% in AUROC among the 21 domains\revision{, from 0.8555 to 0.9009.}




%% file: latex/dataset.tex
\section{Dataset}
Existing detectors are often evaluated on datasets such as SQuAD~\citep{rajpurkar2016squad}, XSum~\citep{narayan2018don}, and  Writing Prompts~\citep{fan2018hierarchical}. However, these datasets typically represent a narrow subset of available data, both in terms of timeliness and domain coverage. This limitation raises concerns about over-fitting and uncertainty regarding how these detectors would perform when deployed in real-world scenarios, highlighting the necessity in creating a dataset of diversely-distributed texts for training.

\subsection{Data Collection}
To ensure the robustness and generalizability of our detection model, we construct a dataset consisting of human-written text from 21 distinct domains, including finance, entertainment, cuisine, etc. For each domain, we collect the texts either by crawling online platforms like Wikipedia or by sampling from publicly available datasets. From these collections, we randomly select 200 \revision{complete paragraphs as text snippets which yields an average length of 120 words among the samples}. For each of these 200 human-written samples per domain, we generate four AI-written counterparts using four state-of-the-art LLMs: GPT-4o~\citep{openai_gpt4o}, GPT-3.5-Turbo~\citep{openai_chatgpt}, Gemini 1.5 Pro~\citep{reid2024gemini}, and Llama-3-70B-Instruct~\citep{meta_Llama3}. This results in a total of 21,000 text samples across all domains. Details on data generation are in Table \ref{tab:dataset_comparison}, and descriptions of the domains and their sources are provided in \S\ref{sec:domains}.

\subsection{Prompt Diversity}
Conventionally, AI-generated text is created by prompting LLMs to either rewrite a given text or continue writing from a given prefix, often using a single, static prompt for the entire process~\citep{mitchell2023detectgpt, bao2023fast, verma2023ghostbuster, mao2024raidar}. However, real-world text generation involves a wide variety of prompts, which can significantly alter the distribution of the generated text. Previous work~\citep{mao2024raidar} has shown that one straightforward way to bypass the \base detector is by using the prompt \texttt{"Help me rephrase it, so that another GPT rewriting will cause a lot of modifications,"} which suggests that data generated by different prompts are different in distribution, indicating the importance of prompt diversity. To address this, we curate a dataset of 200 rewrite prompts, each containing slight variations in phrasing and instructions. For each generated text, a prompt is randomly sampled from this dataset. Examples of the prompts we use are provided below:
\begin{denseitemize}
    \item Refine this for me please:
    \item Please rewrite this content in your own words:
    \item Make this text more formal and professional:
    \item Make this text more casual and friendly:
    \item Rephrase this text in a more elaborate way:
    \item Reframe this content in a more creative way:
    \item Rewrite this text to emphasize the key points:
    \item Help me rephrase it, so that another GPT rewriting will cause a lot of modifications:
\end{denseitemize}
For Gemini rewrite, training on diversely-prompted dataset increases testing AUROC from 0.7302 to 0.7566. For Llama rewrite, AUROC increases from 0.7888 to 0.7970. This shows that diverse prompts enables the model to better capture the distribution of AI texts in the real world, whose generation prompts are expected to vary significantly.

\subsection{Data Cleaning}
In collecting human-written text, we ensure that no data is generated after November 30, 2022, the release date of ChatGPT~\citep{openai_chatgpt}, avoiding contamination of human dataset with AI-generated content. Instead of manually crafting the length, we split all texts into natural paragraphs, yielding an overall average length of 120 words with a standard deviation of 108 words. For AI-generated text, we carefully remove any extraneous suffixes, such as \texttt{“Sure, here is a...,”} to avoid them be detected in this way.

%% file: latex/evaluation.tex
\section{Evaluation}
This section answers the following questions:
\begin{denseitemize}
\item[{\bf Q1:}] How does \name compare with other detectors? (\S\ref{sec:other_classifer_comparison})
\item[{\bf Q2:}] How does \name perform when \ood? (\S\ref{sec:ood_dataset})
\item[{\bf Q3:}] How does \name perform under adversarial attacks? (\S\ref{sec:adv})
\item[{\bf Q4:}] How does \name's training objective compare with directly training for binary classification? (\S\ref{sec:llama_logits})
\item[{\revision{\bf Q5:}}] \revision{How does training on our proposed dataset contribute to \name's performance? (\S\ref{sec:dataset})}
\end{denseitemize}

\subsection{Experiment Setup}
We perform all experiments on one NVIDIA A100 GPU with 40GB RAM. We use 'meta-Llama/Meta-Llama-3-8B-Instruct'~\citep{Llama3modelcard} as the open-sourced rewrite model in all experiments. To fine-tune the Llama model with 8B parameters, we employ 4-bit QLoRA~\citep{dettmers2024qlora}, with parameter \texttt{r} set to 16, \texttt{lora\_alpha} set to 32, and \texttt{lora\_dropout} set to 0.05, unless otherwise noted. We use an initial learning rate of 5e-6, a weight decay of 0.01, and a batch size of 32 to train until convergence. We set the sampling temperature to 0 when using Llama for rewriting during training and detection for deterministic and reproducible results, therefore taking the results from a single run for the experiments. We use 70\% of the dataset for training and the rest for testing in all experiments. Training on the 21 domains takes around six GPU hours and rewriting a single text of 120 words takes an average of 13.5 seconds.

\subsection{Baselines}
Our baseline detectors consist of Fast-DetectGPT~\citep{bao2023fast}, Ghostbusters~\citep{verma2023ghostbuster}, RAIDAR~\citep{mao2024raidar}, and a custom approach named 'Llama Logits,' which involves training a Llama-3-8B model together with a classifier (same size as RAIDAR and \name) on its logits output to perform naive text classification. For Ghostbuster, RAIDAR and 'Llama Logits', we train and test these detectors on the identical training and testing sets as \name. For Fast-DetectGPT, we use its local version available at~\citet{baoguangsheng_fast_detect_gpt}. For 'Llama Logits,' we train its Llama model using the same LoRA configurations as the rewrite model in \name for a fair comparison. We also experiment on using a close-sourced model, Gemini 1.5 Pro~\citep{reid2024gemini} (referred to as Gemini Rewrite), as the rewrite model for RAIDAR in addition to Llama.

\subsection{Compare \name with Other Detectors}
\label{sec:other_classifer_comparison}
We compare the performance of \name with Fast-DetectGPT, Ghostbusters, and RAIDAR (Llama Rewrite and Gemini Rewrite), by measuring the Area Under the Receiver Operating Characteristic Curve (AUROC) scores. The resulting scores for each domain along with their average and standard deviation can be found in Table~\ref{tab:main}. \name constantly outperforms both configurations of RAIDAR in all domains; outperforms Fast-DetectGPT in 20 of 21 domains by an average of 23.04\% in AUROC; and outperforms Ghostbusters in 20 of 21 domains by an average of 19.56\% in AUROC. \name has a 5.62\% lower AUROC score than Fast-DetectGPT on legal document domain, and a 1.62\% lower AUROC score than Ghostbusters on literature creative writing domain, which might be due to the unique distributions of these domains: legal documents require a more rigorous writing style, while creative writing has a more casual style, thus leaving fewer room for rewrite even for human writers. 

In general, the fluctuating AUROC scores indicate the challenging nature of our dataset and the diversity and independence of the distributions across domains. These results also show that \name has better knowledge of the intricate differences between human and AI texts in various domains compared with the baselines, and is more capable in the real-world setting.

\begin{table*}[ht]
    \centering
    \resizebox{\textwidth}{!}{
    \begin{tabular}{lccccc}
    \midrule
    Domain & Fast-DetectGPT & Ghostbusters & \makecell{\revision{RAIDAR} \\ \revision{(Gemini Rewrite)}} & \makecell{\revision{RAIDAR} \\ \revision{(Llama Rewrite)}} & Llama L2R \\
    \midrule
    AcademicResearch & 0.4664 & 0.6597 & 0.7911 & 0.8311 & \textbf{0.8406} \\
    ArtCulture & 0.6292 & 0.6781 & 0.7711 & 0.6750 & \textbf{0.8328} \\
    Business & 0.6829 & 0.8331 & 0.8153 & 0.8369 & \textbf{0.9156} \\
    Code & 0.6808 & 0.3770 & 0.5670 & 0.3840 & \textbf{0.8383} \\
    EducationalMaterial & 0.7474 & 0.8506 & 0.9339 & \textbf{0.9675} & 0.9644 \\
    Entertainment & 0.8392 & 0.8600 & 0.7836 & 0.8319 & \textbf{0.9494} \\
    Environmental & 0.8382 & 0.8447 & 0.9081 & 0.9228 & \textbf{0.9786} \\
    Finance & 0.6879 & 0.7828 & 0.6917 & 0.8153 & \textbf{0.9400} \\
    FoodCuisine & 0.7425 & 0.6703 & 0.7181 & 0.7831 & \textbf{0.9547} \\
    GovernmentPublic & 0.7100 & 0.6833 & 0.7375 & 0.7619 & \textbf{0.8675} \\
    LegalDocument & \textbf{0.8365} & 0.5453 & 0.5528 & 0.6594 & 0.7803 \\
    LiteratureCreativeWriting & 0.7928 & \textbf{0.9456} & 0.8056 & 0.9161 & 0.9294 \\
    MedicalText & 0.5693 & 0.6242 & 0.7614 & 0.7700 & \textbf{0.7857} \\
    NewsArticle & 0.5808 & 0.6800 & 0.7714 & 0.8547 & \textbf{0.9242} \\
    OnlineContent & 0.6292 & 0.5922 & 0.7408 & 0.8231 & \textbf{0.8881} \\
    PersonalCommunication & 0.5392 & 0.7042 & 0.6783 & 0.7233 & \textbf{0.8239} \\
    ProductReview & 0.6467 & 0.7364 & 0.7150 & 0.8075 & \textbf{0.9689} \\
    Religious & 0.6314 & 0.6111 & 0.7772 & 0.8397 & \textbf{0.9775} \\
    Sports & 0.6015 & 0.6561 & 0.6917 & 0.7869 & \textbf{0.8742} \\
    TechnicalWriting & 0.6075 & 0.7242 & 0.8269 & 0.8575 & \textbf{0.9369} \\
    TravelTourism & 0.6210 & 0.7517 & 0.8492 & 0.8897 & \textbf{0.9475} \\
    \midrule
    AVERAGE & 0.6705 & 0.7053 & 0.7566 & 0.7970 & \textbf{0.9009} \\
    STD & 0.1015 & 0.1259 & 0.0928 & 0.1212 & \textbf{0.0634} \\
    \midrule
    \end{tabular}
    }
    \caption{Comparison of detection performance measured with AUROC scores. For Ghostbuster and all rewrite-based detectors, we train a single classifier on the training set of all domains, then test the model's performance on the test set of each individual domain. \textbf{AVERAGE} measures the average performance for all independent domains, and \textbf{STD} measures the standard deviation across domains.}
    \label{tab:main}
\end{table*}

\subsection{\ood Dataset Evaluation}
\label{sec:ood_dataset}
We showed that \name outperforms the state-of-the-art detectors \id in terms of AUROC scores, but it is equally important to assess its robustness under \ood conditions, as training-based detectors are prone to overfitting to familiar domains and generator models. We first evaluate this by showing its performance on \ood datasets.

To assess \name's performance on \ood data, we adopt the M4 dataset~\citep{wang-etal-2024-m4}, \revision{an \ood dataset that is different from our training data in multiple dimensions, including data generation models, text length, decoding strategy, and domains. We show a detailed comparison in Table~\ref{tab:m4}.}

\begin{table*}[ht]
\centering
\resizebox{\textwidth}{!}{
\begin{tabular}{lcc}
\toprule
Dataset                  & Ours                                                      & M4                                    \\ \midrule
Generator   & GPT-3.5-Turbo, GPT-4o, Llama-3-70B, Gemini 1.5 Pro    & BLOOMz, ChatGPT, Davinci, Cohere, Dolly V2 \\ 
Text Length              & Mean: 765 chars, STD: 654 chars                           & Mean: 1365 chars, STD: 244 chars               \\ 
Decoding Strategy        & Nucleus Sampling, Temperature = 1, top\_p = 1            & Varies                                         \\ 
Domains                  & 21 domains                                               & 5 Non-Overlapping English domains              \\ \bottomrule
\end{tabular}
}
\caption{Comparison of characteristics of our dataset and M4 dataset, which we use for \ood evaluation.}
\label{tab:m4}
\end{table*}


The results of the \ood evaluation are presented in Table~\ref{tab:ood}. We include both \id and \ood results to highlight the degree of overfitting for each detector. While the Llama Logits method achieves the highest \id AUROC, its \ood result is the lowest, indicating significant overfitting to the training data. Similarly, Ghostbuster shows overfitting with its \ood AUROC being roughly half of its \id performance. The naive rewrite-based approach shows superior robustness compared with these other methods, but \name trained with reduced parameters, i.e. rank \texttt{r} set to 4 and \texttt{lora\_alpha} set to 8, outperforms Llama Rewrite by 3.45\% \id and 4.67\% \ood. This demonstrates that our fine-tuning does not simply overfits the rewrite model to the training data, but enhances its classification performance across diverse distributions.

\begin{figure}
    \centering
    \includegraphics[width=0.9\columnwidth]{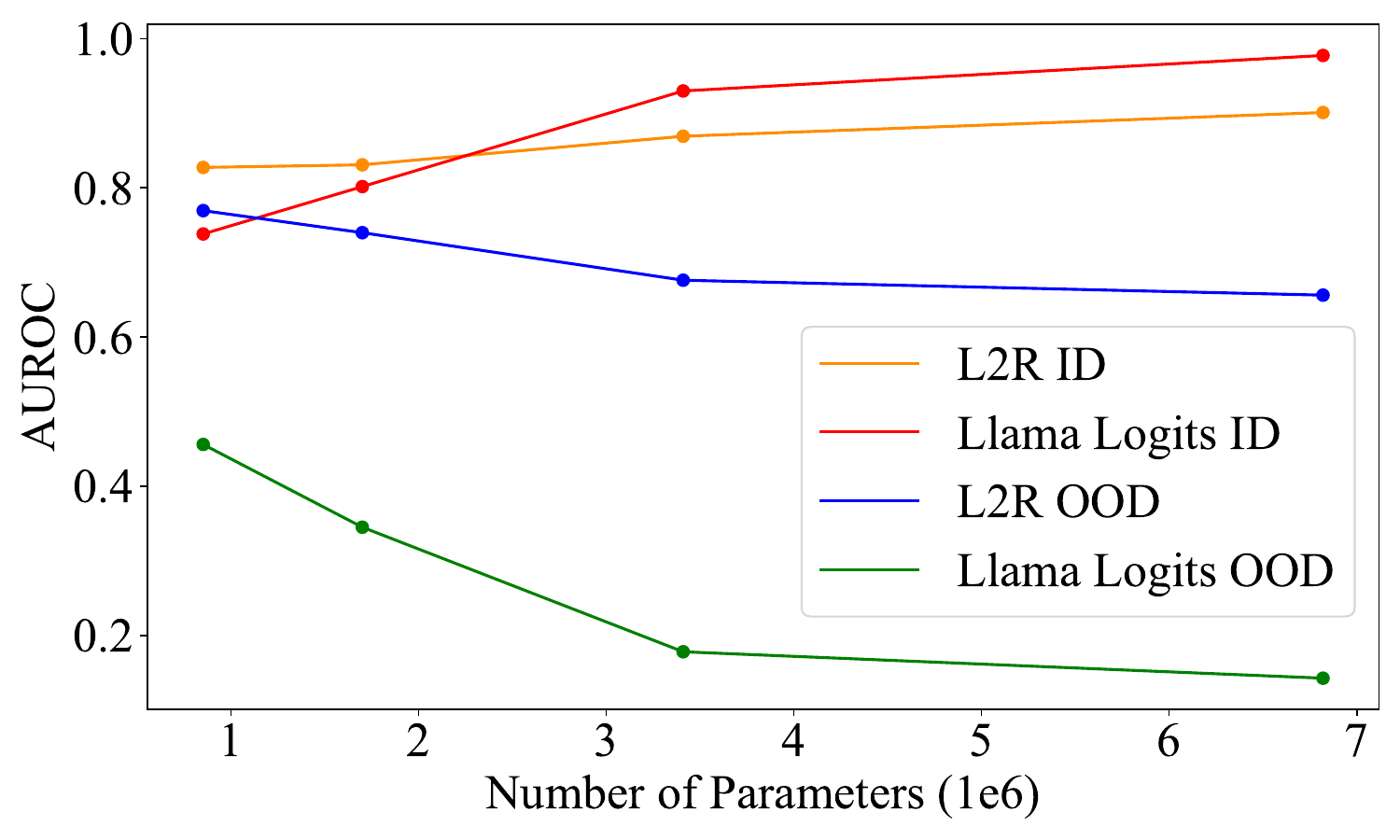}
    \caption{Relationship between the number of trainable parameters and \id and \ood AUROC scores for \name and RAIDAR. As the number of parameters increase from $1\times10^6$ to $7\times10^6$, \revision{both \name and RAIDAR show higher \id performance and lower \ood performance, showing how the effect of overfitting emerges as we increase the LLM's trainable parameters. \name outperforms Llama Logits either \ood or both \id and \ood, showing the superior robustness and accuracy of \name.}}
    \label{fig:plot}
\end{figure}

We notice that reducing the number of training parameters make the model more generalizable, and further investigate the impact of fine-tuning parameters on \name's performance \id and \ood. By adjusting the LoRA parameters \texttt{r} and \texttt{lora\_alpha}, we define four fine-tuning configurations with the number of trainable parameters ranging from 851,968 to 6,815,744, with details listed in \S\ref{sec:lora}. Figure~\ref{fig:plot} illustrates the results, where we observe a consistent increase in \id AUROC, accompanied by a decline in \ood AUROC as the number of parameters grows. This suggests that the model becomes increasingly overfitted to the training distribution. \revision{\name either outperforms Llama Logits \ood or both \id and \ood, and} all four configurations outperform Ghostbusters and Fast-DetectGPT both \id and \ood. Also, the first two configurations surpass RAIDAR in terms of AUROC across both settings.

\subsection{Adversarial Attack} 
\label{sec:adv}
We employ two distinct types of attack to assess \name's robustness against the baseline detectors. For both experiments, we apply the attack to all AI-generated texts in the testing set across all domains, while training \name and the baselines on the unmodified training set and evaluating it on the modified testing set.

\subsubsection{Decoherence Attack} 
\citet{bao2023fast} introduces the decoherence attack where two adjacent, randomly selected words are transposed in all sentences longer than 20 words within a paragraph for AI texts.~\citet{bao2023fast} demonstrated that this simple attack can be highly effective in degrading the performance of sate-of-the-art detectors, without affecting the core meaning of the input. We present the results of this attack in Table~\ref{tab:adv}, where \name achieves the highest AUROC on samples subjected to this attack, indicating its superior robustness compared to other models. This is because our rewrite-based objective function for fine-tuning teaches the model the innate distributions of human and AI texts, instead of relying on brittle statistical features that are easily altered through this simple attack.

\subsubsection{Rewrite Attack} 
\citet{mao2024raidar} introduces the rewrite attack where a GPT-3.5-Turbo model is prompted to refine an input paragraph, generated by AI, in such a way that a subsequent rewrite by another GPT model would result in significant changes.~\citet{mao2024raidar} showed that this type of attack is particularly effective against rewrite-based detectors, as it disrupts the rewrite we use for classification. As shown in Table~\ref{tab:adv}, \name again achieves the highest AUROC on these attack samples, further demonstrating its robustness through fine-tuning. This is because its fine-tuning objective creates separable gap between human and AI rewrite ratios that is large enough so that the attack samples remain in the AI distribution despite the perturbations. Concretely, the average edit ratio of human texts is 0.6981, and of AI texts is 0.8606. After attack, the ratio for AI decreases to 0.8386, which suggests that the rewrite attack is effective in shifting the AI distribution towards human, but there still exists a clear gap between both distributions, so that \name's classification performance only degrades marginally.

\begin{table}[t!]
\centering
\resizebox{\columnwidth}{!}{
\begin{tabular}{lcc}
\toprule
Model & In-Distribution & Out-of-Distribution \\
\midrule
Ghostbusters               & 0.7053 & 0.3888 \\
Fast-DetectGPT             & 0.6705 & 0.6408 \\
Llama Logits               & \textbf{0.9774} & 0.1426 \\
Llama Logits (Reduced Params)  & 0.8016 & 0.3450 \\
Llama Rewrite              & 0.7970 & 0.6931 \\
\midrule
Llama L2R                  & 0.9009 & 0.6561 \\
Llama L2R (Reduced Params)  & 0.8315 & \textbf{0.7398} \\
\bottomrule
\end{tabular}
}
\caption{\id and \ood performance measured in AUROC scores. For \name and Llama logits, the "Reduced Params" models are tuned with approximately 1/4 of the parameters for better generalizability. With reduced parameters, \name has the highest \ood AUROC, outperforming the naive Llama rewrite both \id and \ood by 3.45\% and 4.67\%, respectively, suggesting its generalizability through fine-tuning.}
\label{tab:ood}
\end{table}

\begin{table}[t!]
\centering
\resizebox{\columnwidth}{!}{
\begin{tabular}{lccc}
\toprule
Model & No Attack & Decoherence Attack & Rewrite Attack \\
\midrule
Ghostbusters     & 0.7053 & 0.4730 & 0.4061 \\
Fast-DetectGPT   & 0.6705 & 0.4984 & 0.5100 \\
Llama Logits     & \textbf{0.9774} & 0.7281 & 0.6563 \\
Llama Rewrite    & 0.7970 & 0.7681 & 0.7944 \\
\midrule
Llama L2R        & 0.9009 & \textbf{0.8746} & \textbf{0.8927} \\
\bottomrule
\end{tabular}
}
\caption{Adversarial attack results. While all detectors show performance degredation under attack, \name has the highest AUROC in both setting, suggesting its robustness through fine-tuning.}
\label{tab:adv}
\end{table}

\subsection{Compare \name with Direct Fine-Tuning}
\label{sec:llama_logits}
A valid concern regarding \name's superior performance is whether it is due to our fine-tuning objective, which enhances model's rewriting ability, or is solely from the fact that we exploit the vast parameters of an LLM. To answer this question, we compare \name with the 'Llama Logits' baseline in Table~\ref{tab:ood} and \ref{tab:adv}. The Llama logits detector involves fine-tuning a Llama-3-8B model not for rewrite, but directly for binary classification. 

In \S\ref{sec:ood_dataset}, we show that despite the Llama classifier has the highest \id AUROC score among all detectors, surpassing \name by 7.65\%, it has the lowest AUROC when evaluated \ood, up to 51.35\% lower than \name, which suggests that its performance \id is due to overfitting. This highlights the importance of our fine-tuning objective function in ensuring domain-agnostic detection accuracy. Also, the Llama classifier is inferior under adversarial attacks, with 14.65\% and 23.64\% lower AUROC for decoherence and rewrite attacks, respectively. This again shows \name's robustness in capturing the true underlying distributions of human and AI data.

\subsection{Effectiveness of the Diverse Dataset}
\label{sec:dataset}
While there exists public datasets that emphasize data diversity, including RAID~\citep{dugan-etal-2024-raid}, RuTAD~\citep{ruatd}, and MAGE~\citep{li2024mage}, the contribution of our proposed dataset lies in its ability to train a robust and generalizable L2R model. We show this by training \name on MAGE using the same number of texts and under the same configurations, then test its performance \id and \ood on the M4 dataset. We compare the results in \ref{tab:dataset_comparison}, where \name trained on our dataset has 15.98\% higher \ood AUROC, suggesting that the diverse text distributions in our dataset is effective in training a robust and generalizable \name model.

\begin{table}[h!]
\centering
\resizebox{\columnwidth}{!}{
\begin{tabular}{lcc}
\toprule
Training Dataset & ID AUROC & OOD AUROC \\
\midrule
MAGE & 0.8333 & 0.4963 \\
Ours & \textbf{0.9009} & \textbf{0.6561} \\
\bottomrule
\end{tabular}
}
\caption{Comparison of \name's \id and \ood performance when trained on MAGE and our dataset. The superior OOD performance when trained on our dataset suggests its effectiveness.}
\label{tab:dataset_comparison}
\end{table}

%% file: latex/conclusion.tex
\section{Conclusion}
We present \name, a method designed to enhance the detection of LLM-generated text by learning to rewrite more on LLM-generated inputs and less on human generated inputs. \name excels in identifying LLM-generated content collected across various models and 21 unique domains, both \id and \ood, and under adversarial attacks. Our work demonstrates that LLMs can be trained to detect content generated by other LLMs, surpassing previous detection methods in accuracy and generalizability.

%% file: latex/limitations.tex
\section{Limitations}
A limitation of ours is the relatively slow inference runtime. As most zero-shot detectors only requires a forward pass from the LLM being used, we need to call generate to create a rewrite. Nevertheless, this problem would be well alleviated considering the rapid enhancement in LLM efficiency and computing power.

%% file: latex/appendix.tex
\newpage
\section{Appendix}
\subsection{Dataset Details}
\label{sec:domains}
Our dataset encompasses 21 indepedent English domains. Table~\ref{tab:datasets} shows the source and license for each domain. For all domains, we manually verify that no personal or offensive content are included. For domains that are taken from third-party datasets, we use the data consistent with their intended use (detection of machine generated text).

\begin{table*}[h]
    \centering
    \resizebox{\textwidth}{!}{
    \begin{tabular}{lll}
        \toprule
        \textbf{Category} & \textbf{Source} & \textbf{License} \\
        \midrule
        AcademicResearch & Arxiv abstracts~\citep{mao2024raidar} & Various CC licenses \\
        ArtCulture & Wikipedia & CC BY-SA \\
        Business & Wikipedia & CC BY-SA \\
        Code & Code snippets~\citep{mao2024raidar} & MIT \\
        EducationalMaterial & Ghostbuster essays~\citep{verma2023ghostbuster} & CC BY 3.0 \\
        Entertainment & IMDb dataset~\citep{imdb_dataset}, Stanford SST2~\citep{socher-etal-2013-recursive} & IMDb terms of use, CC Zero \\
        Environmental & Climate-Ins~\citep{spokoyny2023towards} & CC Zero \\
        Finance & Hugging Face FIQA~\citep{thakur2021beir} & CC BY-NC \\
        FoodCuisine & Kaggle fine food reviews~\citep{mcauley2013amateurs} & CC Zero \\
        GovernmentPublic & Wikipedia & CC BY-SA \\
        LegalDocument & CaseHOLD~\citep{zheng2021does} & Apache 2.0 \\
        CreativeWriting & Writing Prompts~\citep{fan2018hierarchical} & MIT \\
        MedicalText & PubMedQA~\citep{jin2019pubmedqa} & MIT \\
        NewsArticle & XSum~\citep{narayan2018don} & MIT \\
        OnlineContent & Hugging Face blog authorship~\citep{schler2006effects} & Non-commercial \\
        PersonalCommunication & Hugging Face daily dialogue~\citep{li2017dailydialog} & CC-BY-NC-SA 4.0 \\
        ProductReview & Yelp reviews~\citep{mao2024raidar} & Yelp terms of use \\
        Religious & Bible, Buddha, Koran, Meditation, and Mormon & N/A \\
        Sports & Olympics website~\citep{olympics} & Olympics terms of use \\
        TechnicalWriting & Scientific articles~\citep{mosca-etal-2023-distinguishing} & CC Zero \\
        TravelTourism & Wikipedia & CC BY-SA \\
        \bottomrule
    \end{tabular}
    }
    \caption{Source and license for each domain in our dataset.}
    \label{tab:datasets}
\end{table*}

\subsection{Generation Prompts}
\label{sec:prompts}
Our dataset encompasses 200 different prompts for generating AI data. Here is an incomplete list of the prompts we used:
\begin{denseitemize}
    \item Refine this for me please:
    \item Please rewrite this content in your own words:
    \item Make this text more formal and professional:
    \item Make this text more casual and friendly:
    \item Rephrase this text in a more elaborate way:
    \item Reframe this content in a more creative way:
    \item Can you make this sound more enthusiastic?
    \item Rewrite this passage to emphasize the key points:
    \item Help me rephrase it, so that another GPT rewriting will cause a lot of modifications:
\end{denseitemize}

\subsection{Effectiveness of the Diverse Prompt in Data Preparation}
\label{sec:diverse_eval}
Our  dataset involves 21 independent domains, four source LLMs, and 200 generation prompts, resembling real-world use cases for text detectors compared with traditional evaluation datasets which are usually constrained to one single domain and generation prompt. To prove the superiority of our dataset in training more capable detection models, we create a parallel nondiverse dataset which is created on the same number of domains and source LLMs, but generate the AI data with only with the prompt \texttt{"Rewrite this for me please."} Then, we train two RAIDAR detectors without fine-tuning, on the non-diverse dataset, and evaluate it on the diverse dataset. As shown in Table~\ref{tab:ablation1}, the diverse prompts yields to 2.64\% increase in AUROC score if the rewrite model is Gemini 1.5 Pro, and 0.82\% increase in AUROC score if the rewrite model is Llama-3 8B. This validates the effectiveness of the diverse prompts we were using, and suggests that such diversity could help the detector to capture more information about real world data distributions.

\begin{table}[ht!]
    \centering
    \resizebox{\columnwidth}{!}{
    \begin{tabular}{lcccc}
        \toprule
        Dataset & Rewrite Model & AUROC \\
        \midrule
        Single-Prompt & Gemini & 0.7302 \\
        Multi-Domain Dataset & Llama & 0.7888 \\
        \midrule
        Multi-Prompt & Gemini & \textbf{0.7566} \\
        Multi-Domain Dataset & Llama & \textbf{0.7970} \\
        \bottomrule
    \end{tabular}
    }
    \caption{Comparison of AUROC scores for Gemini and Llama rewrite models on nondiverse and duverse Datasets. Diverse prompting in the training set enhances detection performance for both models.}
    \label{tab:ablation1}
\end{table}

\subsection{LoRA Configurations for Fine-Tuning}
Table~\ref{sec:lora} lists the four fine-tuning configurations we use in \S\ref{sec:ood_dataset}.
\label{sec:lora}

\begin{table}[h!]
\centering
\resizebox{0.4\textwidth}{!}{
\begin{tabular}{ccc}
\midrule
\texttt{r} & \texttt{lora\_alpha} & \text{Trainable Parameters} \\
\midrule
2  & 4  & 851,968 \\
4  & 8  & 1,703,936 \\
8  & 16 & 3,407,872 \\
16 & 32 & 6,815,744 \\
\midrule
\end{tabular}
}
\caption{Parameter settings for LoRA fine-tuning.}
\end{table}

\subsection{Effectiveness of the Calibration Loss}
\label{sec:loss_eval}
\begin{figure}[ht!]
    \centering
    \includegraphics[width=\columnwidth]{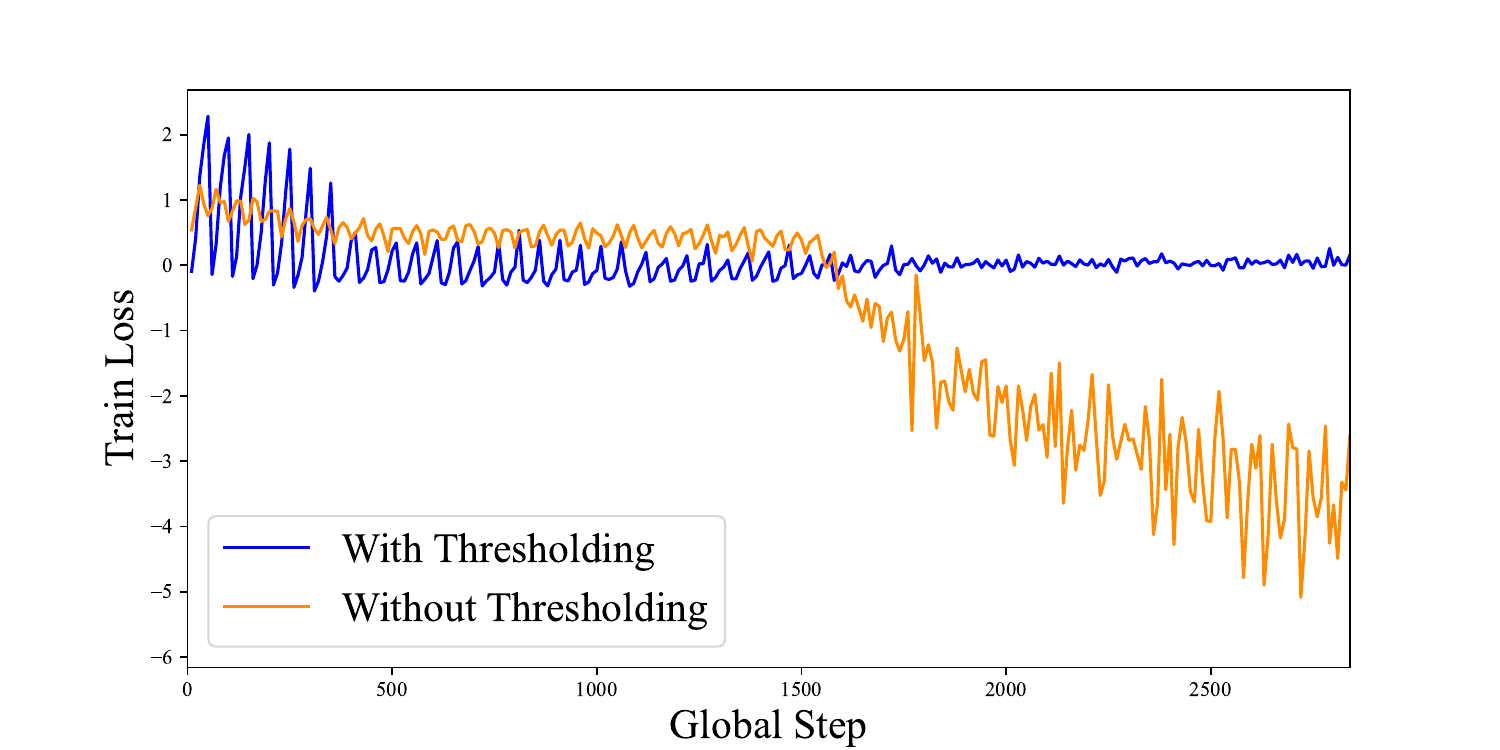}
    \caption{Training loss curves for the rewrite model. The \textcolor{orange}{orange} plots the loss trained without the calibration method, and the \textcolor{blue}{blue} line plots the loss trained with the method. The later one exhibits faster convergence and higher stability than the former one.}
    \label{fig:loss}
\end{figure}

An important contribution of ours that improves the fine-tuning performance is the calibration loss, as proposed in \S3.4. Without this loss, the model tends to overfit during fine-tuning as shown in Figure~\ref{fig:loss}, where the model loss drastically decrease after 1500 steps, resulting in verbose rewrite even for LLM-generated text. We conduct an ablation study on five domains where the AUROC score is only 0.62 after the model overfits. We hypothesized that this technique could benefit model learning because the threshold effectively prevents further modification to model weights once an input, labeled either AI or human, falls in its respectively distribution already. Since our purpose is simply to draw a boundary rather than separate the distributions as much as possible, this halt in further weight adjustments facilitates the model to only perform parameter update on those inputs which are not yet correctly classified, so that it could converge more efficiently and effectively. Concretely, applying the calibration loss improves detection performance by 4.54\% in AUROC among the 21 domains, even comparing with a model tuned with the loss before over-fitting.

\subsection{Different Ways to Generate \ood Data}
\label{sec:ood_data}
There exists a variety of ways to generate \ood data, including using different generation models, decoding strategies, text lengths, and writing styles. While we show how M4, the \ood dataset we use for evaluation, is distinct from our training domain in all above aspects in \ref{tab:m4}, we conduct an additional ablation study on how different text length and decoding strategy alone could influence detection performance in Table \ref{tab:ablation2}

\begin{table*}[h!]
\centering
\begin{tabular}{llccc}
\toprule
Avg Length & Decoding Strategy & Fast-DetectGPT & RAIDAR & L2R \\ 
\midrule
120                 & Nucleus Sampling                & 0.6833                  & 0.8186          & 0.9213       \\
60                  & Nucleus Sampling                & 0.6500                  & 0.7635          & 0.8632       \\
120                 & Greedy Decoding \& Beam Search  & 0.6897                  & 0.8009          & 0.8750       \\ 
\bottomrule
\end{tabular}
\caption{Performance comparison of different setups across models.}
\label{tab:ablation2}
\end{table*}

We use 200 randomly selected texts from our dataset for both studies. For decoding strategy, we use greedy decoding for GPT and Gemini models and beam search with num\_beams=5 for the Llama model. For text length, we chunk the texts to an average length of 60. We test L2R on the two datasets and show results below, where L2R outperforms RAIDAR by 9.97\% for the length ablation and 7.41\% for the decoding strategy ablation in AUROC. This further shows \name's robustness to different \ood data distributions.


\subsection{Rewrite Examples}

\label{sec:examples}
\begin{figure*}[ht!]
    \centering
    \includegraphics[width=\textwidth]{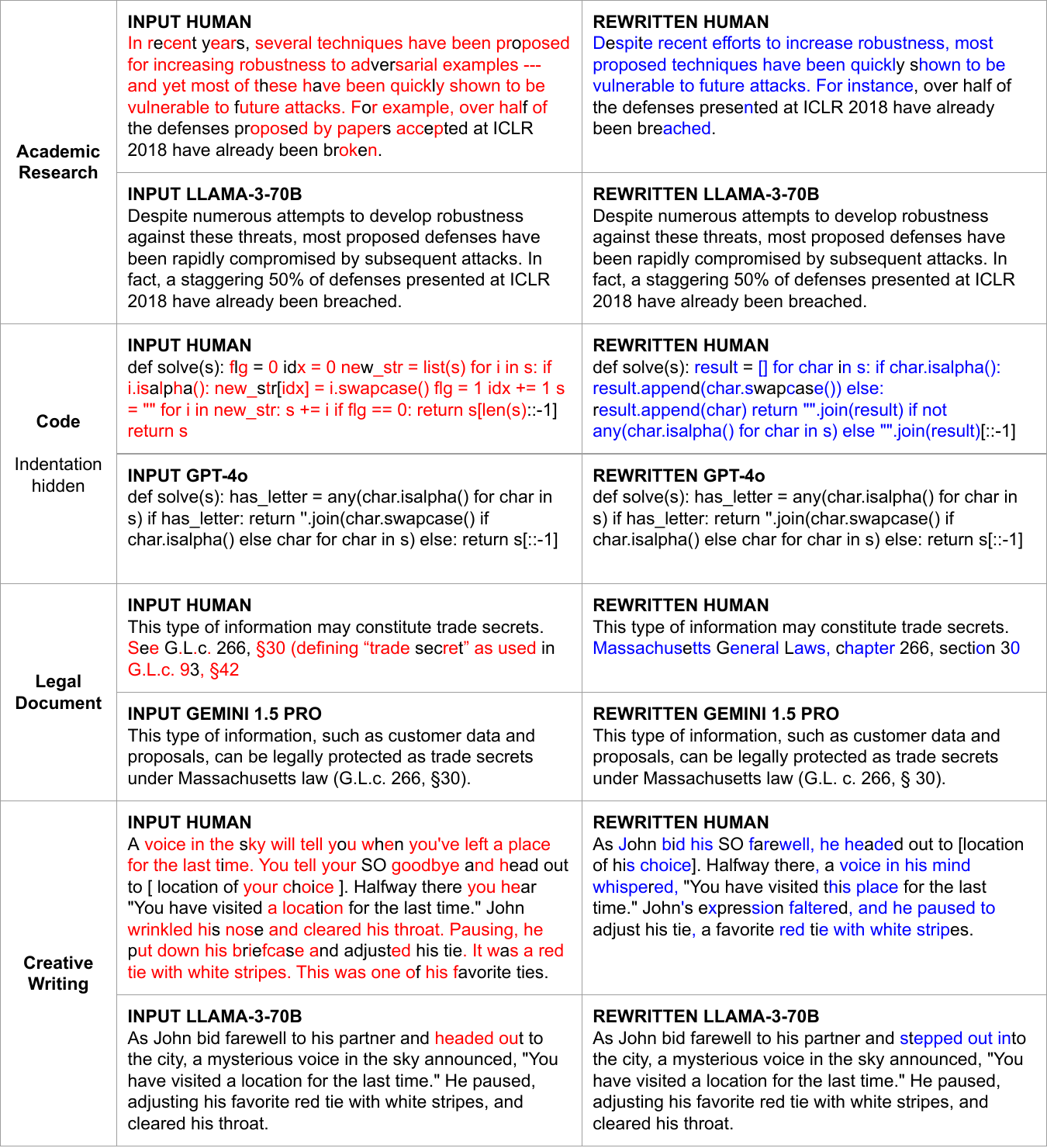}
    \caption{Examples of texts in our proposed dataset along with the amount of edits L2R model gives for human and LLM data. Deleted characters are marked in \textcolor{red}{red}, inserted characters are in \textcolor{blue}{blue}, and unmodified characters are in \textbf{black}. The examples demonstrate the diverse domains and source LLMs available in the dataset, as well as \name's ability in separating human and LLM texts via rewriting. }
    \label{fig:dataset}
\end{figure*}